\documentclass[10pt,twocolumn,letterpaper]{article}

\usepackage{cvpr}
\usepackage{times}
\usepackage{epsfig}
\usepackage{graphicx}
\usepackage{amsmath}
\usepackage{amssymb}
\usepackage{subfigure}
\usepackage{mathrsfs}
\usepackage{latexsym}
\usepackage{float}

\usepackage[pagebackref=true,breaklinks=true,letterpaper=true,colorlinks,bookmarks=false]{hyperref}

\cvprfinalcopy

\ifcvprfinal\fi

\begin{document}
\title{TSGCNet: Discriminative Geometric Feature Learning with Two-Stream Graph Convolutional Network for 3D Dental Model Segmentation}

\author{\hspace*{-23pt}Lingming Zhang\textsuperscript{\rm 1 }\thanks{Equal contribution.} \qquad 
Yue Zhao\textsuperscript{\rm 1 }\footnotemark[1] \qquad 
Deyu Meng\textsuperscript{\rm 2 } \qquad 
Zhiming Cui\textsuperscript{\rm 3 } \qquad 
Chenqiang Gao\textsuperscript{\rm 1 }\thanks{Corresponding author.}\qquad 
Xinbo Gao\textsuperscript{\rm 1 } \\ \qquad 
Chunfeng Lian\textsuperscript{\rm 2 } \qquad 
Dinggang Shen\textsuperscript{\rm 4 5 6 }\footnotemark[2]\\
    $^1$Chongqing University of Posts and Telecommunications, Chongqing, China \\ 
    $^2$Xi'an Jiaotong University, Xi'an, China\\
    $^3$The University of Hong Kong\\
    $^4$School of Biomedical Engineering, ShanghaiTech University, Shanghai, China\\
    $^5$Shanghai United Imaging Intelligence Co., Ltd., Shanghai, China\\
    $^6$Department of Artificial Intelligence, Korea University, Seoul 02841, Republic of Korea\\
}

\maketitle

\begin{abstract}
The ability to segment teeth precisely from digitized 3D dental models is an essential task in computer-aided orthodontic surgical planning. 
To date, deep learning based methods have been popularly used to handle this task. 
State-of-the-art methods directly concatenate the raw attributes of 3D inputs, namely coordinates and normal vectors of mesh cells, to train a single-stream network for fully-automated tooth segmentation. This, however, has the drawback of ignoring the different geometric meanings provided by those raw attributes. This issue might possibly confuse the network in learning discriminative geometric features and result in many isolated false predictions on the dental model.
Against this issue, we propose a two-stream graph convolutional network (TSGCNet) to learn multi-view geometric information from different geometric attributes. 
Our TSGCNet adopts two graph-learning streams, designed in an input-aware fashion, to extract more discriminative high-level geometric representations from coordinates and normal vectors, respectively. 
These feature representations learned from the designed two different streams are further fused to integrate the multi-view complementary information for the cell-wise dense prediction task.
We evaluate our proposed TSGCNet on a real-patient dataset of dental models acquired by 3D intraoral scanners, and experimental results demonstrate that our method significantly outperforms state-of-the-art methods for 3D shape segmentation.
\end{abstract}


\begin{figure}[t]
\begin{center}
\includegraphics[scale=0.85]{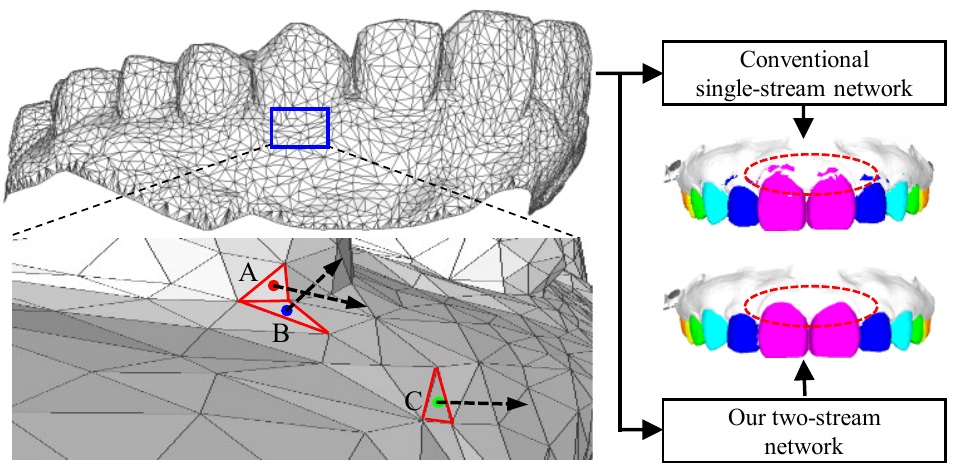}
\end{center}
   \caption{An illustration of 3D dental model. In the local space indicated by the blue box, cell A and cell B are spatially close but with much different normal vectors (indicated by the black arrows in the zoomed view). In contrast, cell A and cell C have similar normal vectors but are far from each other. It suggests that coordinates and normal vectors provide completely different geometric information. Hence, simply concatenating them as a single feature vector (commonly used in the conventional single-stream networks) cannot properly integrate such complementary information to learn more discriminative geometric representations for cell classification, which will result in many isolated false predictions on dental model (as indicated by one of red dotted circle).}
\label{normal}
\vspace{-0.5em}
\end{figure}

\section{Introduction}
An essential task in computer-aided-design system for orthodontic treatment is to provide accurate segmentation of teeth on digitalized 3D dental models reconstructed by intraoral scanners (IOS).
This segmentation information can be used for aiding clinical diagnose, providing digital teeth shape information for personal surgical-orthodontic planning, quantifying the difference between expected and clinical treatment results to adjust orthodontic treatment plan, etc.
 However, segmenting each tooth from the gingiva is practically challenging, mainly due to heterogeneous tooth appearance:
 \textbf{i)} Although most human teeth have common geometric characteristics, their shapes are unique and vary dramatically across individuals. 
 \textbf{ii)} Orthodontic patients usually have atypical conditions such as missing, crowded and/or misaligned teeth, all of which may produce indistinct tooth boundaries.
 \textbf{iii)} Noise and occlusion during scanning may result in a partially reconstructed dental surface with missing parts. 

To deal with these challenges, various (semi-) automated methods have been proposed for tooth segmentation on 3D dental models. 
Conventional approaches typically perform segmentation by using pre-selected geometric properties (e.g., the 3D coordinates, normal vectors and curvature)~\cite{T1,T2,T3,T4,T5,T6,T7,T8} or projecting 3D meshes onto 2D images~\cite{T9,T10}.
Due to the requirement of manual initialization, the efficacy of such semi-automated methods relies on the professional knowledge and experience.
Furthermore, the robustness of these conventional methods may be hampered since the exclusive use of low-level geometric properties would not be able to segment teeth with extreme appearances accurately. 

Recently, deep learning-based methods have been proposed to learn task-oriented feature representations for fully-automated tooth segmentation. 
Some of these methods \cite{Dental1, dental2} transformed mesh vertices/cells as ordered 2D image-like (or volumetric) inputs and then applied general convolutional neural networks (CNNs) to perform segmentation.  
Although straightforward, such operations tend to ignore the unordered nature of geometric data. They also incline to introduce additional computational costs and quantization errors during the potential voxelization stage. 
To avoid additional data pre-processing, more recent methods~\cite{dental3, dental4, dental5} applied or extended existing point-cloud segmentation networks to perform vertex/cell-wise semantic labeling of 3D dental meshes.
As the network inputs, the 3D coordinates and normal vectors (of mesh vertices/cells) are typically concatenated in these methods to train a single-stream network.
However, considering that the coordinate indicates the cell spatial position, while the normal vector represents the cell morphological structure, directly combing these two completely different attributes as a single feature vector tends to weaken their geometric discrimination (e.g., an example is shown in Fig.~\ref{normal}). Hence, this would confuse those conventional single-stream networks in learning discriminative geometric features for cell classification, potentially resulting in isolated false predictions on the dental model.

To resolve these issues, we propose a two-stream graph convolutional network (TSGCNet) in this paper to learn multi-view geometric information for end-to-end tooth segmentation on 3D dental models.
In order to eliminate the mutual confusion caused by mixed geometric inputs, our TSGCNet starts with two parallel branches, namely C-stream and N-stream, to learn independently multi-scale feature representations from coordinates and normal vectors, respectively. 
Besides, considering different geometric meanings of those attributes, the two streams are also constructed by different graph-learning strategies designed in an input-aware fashion. 
That is, C-stream adopts graph-attention convolutions~\cite{Att2} to learn the coarse structures of different teeth from coordinates, while the N-stream adopts graph max-pooling to extract distinctive structural details~\cite{Att2} from the normal vectors, which can further help C-stream to distinguish neighboring cells belonging to different classes (e.g., boundaries between adjacent teeth or between teeth and gingiva).
These multi-scale geometric representations produced by the two parallel streams are further fused by the subsequent multi-layer perceptrons (MLPs) to learn complementary multi-view information for dense labeling of all cells on the mesh surface.

The main contributions of this paper can be summarized as follows:
\begin{itemize}
\item We propose a novel two-stream  graph convolutional network that can independently process coordinates and normal vectors to learn more discriminative geometric features for 3D dental model segmentation. 
\item We design two different graph-based feature aggregation modules in an input-aware fashion to consume cell coordinates and normal vectors, respectively. That is, the C-stream adopts graph attention convolutions to capture the coarse structure of teeth from coordinates, while the N-stream extract distinctive structural details from normal vectors.
\item Our TSGCNet is evaluated on a clinical dataset of 3D dental models for different orthodontic patients digitized by IOS. The experimental results show that our TSGCNet significantly outperform state-of-the-art 3D shape segmentation methods.
\end{itemize}

\begin{figure*}[t]
\begin{center}
\includegraphics[scale=1.05]{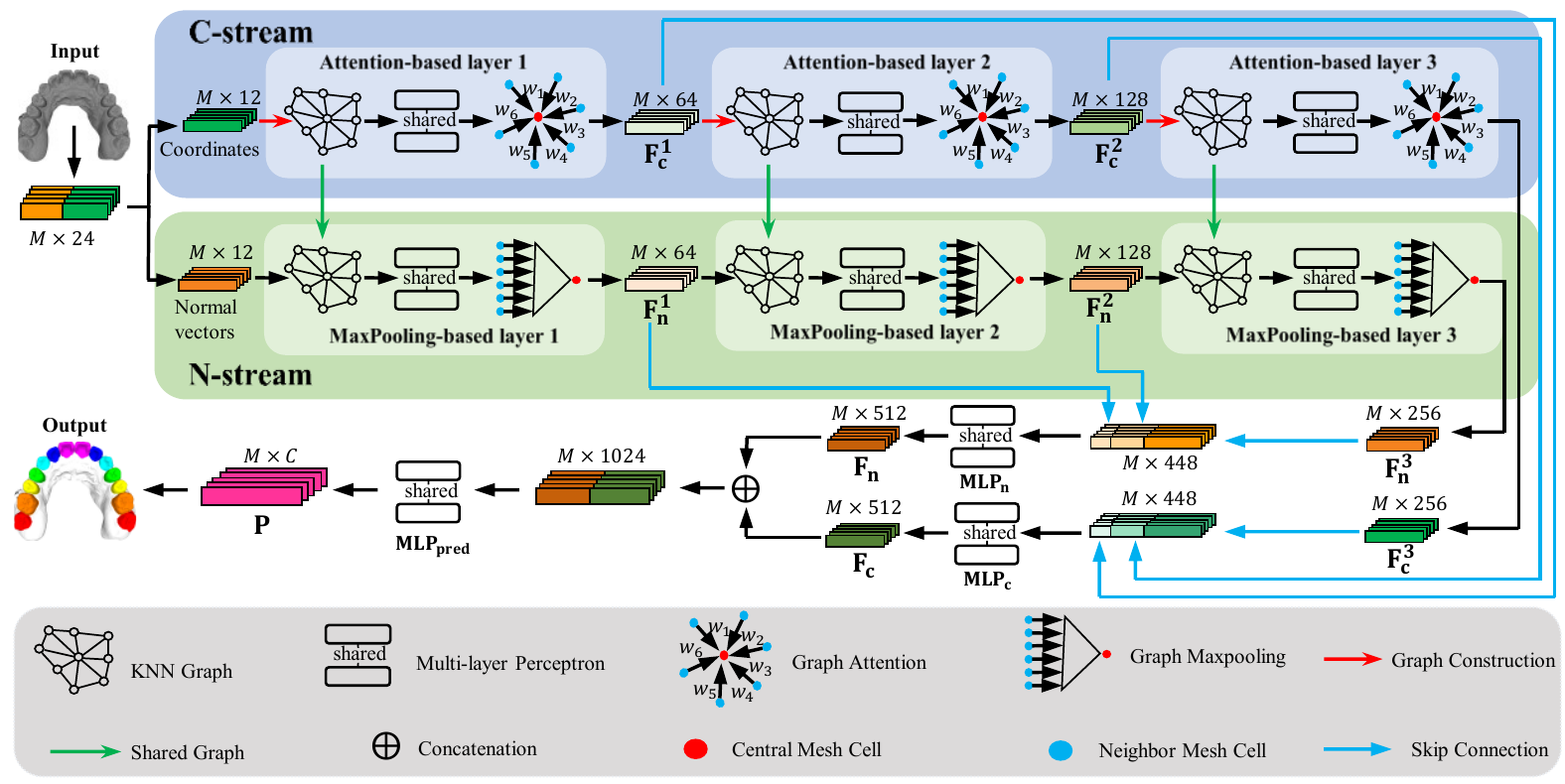}
\end{center}
\caption{Structure of our TSGCNet. The network takes raw mesh data as inputs, and adopts two independent graph convolutional streams(i.e., C-strean and N-stream) to learn discriminative geometric representations from different features(i.e., 3D coordinates and normal vectors of meshes). Then, the high level feature produced by each stream are fused for final mesh-wise tooth segmentation.}
\label{structure}
\end{figure*}

\section{Related Work}
\subsection{3D Shape Segmentation}
Diverse deep learning methods have been proposed for 3D shape classification and segmentation. Some of these approaches voxelized 3D shapes into regular 3D grids~\cite{PV1,PV2,PV3,PV4,PV5,PV6,PV7,PV8} or rendered them into multi-view 2D images~\cite{PM1,PM2,PM3,PM4,PM5}, after which standard CNNs were applied to extract features. 
Such kinds of operations inevitably resulted in spatial information loss and quantization artifacts, inclining to hamper 3D shape segmentation accuracy. 

A pioneering network, PointNet~\cite{Pointnet} consisting of successive MLPs and a symmetric function (e.g., global max-pooling), was proposed to learn directly translation-invariant geometric features from irregular 3D data (e.g., point clouds).
Although achieving promising results in multiple tasks, PointNet ignored the local spatial relationships on 3D shapes as the architecture learns features for each cell independently. 
To address this limitation, PointNet++~\cite{Pointnet++} constructed a hierarchical architecture with successive sampling layers, grouping layers and PointNet modules. 
It explicitly captured and integrated local-to-global spatial information on 3D shapes, achieving better performance than the original PointNet in both the classification and segmentation tasks.
Similarly, to model spatial dependencies of neighboring points, PointCNN~\cite{pointcnn} adopted an encoder-decoder architecture with $\chi$-transformations of unordered points to perform general convolutional operations. 
More recent works further extended PointNet++ by integrating attention modules~\cite{AM1,AM2}, geometry sharing modules~\cite{geometry} and edge branches~\cite{EdgeBranch}.  
Although those PointNet-like methods can directly consume the raw 3D data for end-to-end inference, they usually missed detailed semantic information since the coarse modeling of local dependencies.

Considering that Graph CNNs have shown great success and flexibility in learning from data with irregular structures, some graph-based methods were also proposed for 3D shape recognition and segmentation. 
They usually defined the spatial relations between points/cells as a graph and then used spectral-based~\cite{spectral1,spectral2} or spatial-based~\cite{spatial1} graph convolutions to aggregate local information.
To extract more detailed geometric features, some researchers~\cite{Att1,Att2,Att3} additionally applied attention mechanism during the feature aggregation step.

\subsection{3D Dental Model Segmentation}
Conventional tooth segmentation methods using pre-selected geometric properties can be roughly grouped as curvature-based methods~\cite{T1,T2,T3,T4,T5}, contour-line-based methods~\cite{T6,T7} and harmonic-field-based methods~\cite{T8}. 
Another commonly used strategy is to project the 3D dental models as 2D images, after which the teeth are segmented in 2D and reprojected back to 3D. 
Due to the typical requirement of manual steps and domain knowledge, the efficacy of such semi-automated methods heavily depends on the operator experience.

In recent years, several deep learning-based methods have been proposed for fully-automated tooth segmentation on dental models. 
Typically, Xu \etal.~\cite{dental2} proposed to reshape handcrafted geometric features as 2D image patches to train CNNs for classifying the mesh cells.
Tian \etal.~\cite{Dental1} proposed to voxelize the dental model with a sparse octree partitioning~\cite{PV4}, after which 3D CNNs are applied for tooth segmentation. 
Although those methods using standard CNNs can learn task-oriented feature representations for segmentation, converting the original input into grid format either ignores the unordered nature of the geometric data~\cite{dental2} or may introduce additional quantization errors during the voxelization step~\cite{Dental1}. 
To address this limitation, Zanjani \etal.~\cite{dental3} proposed an end-to-end network that integrates PointCNN~\cite{pointcnn} with a discriminator to directly segment the raw dental surfaces acquired by IOS. 
Lian \etal.~\cite{dental4} extended PointNet~\cite{Pointnet} by adding a multi-scale graph-constrained module to extract fine-grained local geometric features from dental mesh data.
Instead of using solely the 3D coordinates (e.g., in \cite{dental3}), Lian \etal.~\cite{dental4} combined 3D coordinates and normal vectors as the network input to improve the segmentation performance. 

However, since coordinates and normal vectors are completely different geometric meanings of a 3D shape, directly combining these low-level features as a single-stream input would confuse the learning of discriminative geometric representations. Different from those methods, our TSGCNet adopts two graph-learning streams, designed in an input-aware, to independently learn feature representations from coordinates and normal vectors. This can eliminate the mutual confusion caused by mixed geometric inputs and extract.

\section{The Proposed Method}
\subsection{Overview}
Given a 3D dental model with $M$ mesh cells, we define the input of our TSGCNet as a $M\times24$ matrix. That is, each specific cell is described by a $24$-dimensional vector, including the 3D coordinates ($12$ elements) and normal vectors ($12$ elements) of four points (i.e., the cell's three vertices and its central point).
As illustrated in Fig.~\ref{structure}, our TSGCNet starts with a two-stream architecture, which adopts a C-stream and a N-stream to learn more discriminative geometric representations from the coordinates and normal vectors, respectively. 
Thereafter, the features produced by these two complementary streams are further fused to learn higher-level representation for final prediction. 
The output of our TSGCNet is an $M\times C$ matrix, with each row denoting the probabilities of the respective cell belonging to $C$ different classes.

\subsection{Two-Stream Architecture}
\label{Two Stream}
\paragraph{C-Stream.} Our C-stream is designed to learn the basic topology of a dental model. As shown in Fig.~\ref{structure}, given the input of a $M\times 12$ coordinate matrix, a series of graph-attention layers are successively applied in the forward path to extract multi-scale geometric features from the coordinate aspect. 
In each layer of the C-stream, a KNN graph $G$ is first constructed for the $M$ cells in terms of the input features. 
Specifically, for each cell (i.e., a central node), we find its $K$ nearest cells with the smallest Euclidean distance in feature space. 
Let the resulting graph be $G(V,E)$, where $V=\{m_1,m_2,...,m_M\}$ and $E\subseteq| V|\times|V|$ represent the set of nodes (mesh cells) and the set of edges (defined by KNN connectivity), respectively. For each node $m_i\in V$, we denote its KNN as $\mathcal{N}(i)$. 

After building the KNN graph $G$ in each layer, a shared MLP is applied to learn embedded features on each $\mathcal{N}(i)$. Let $\mathbf{f}^{l}_{i}\in \mathbb{R}^d$ (e.g., $d=12$ in the first layer) denote the input feature vector of $m_i$ in the $l$-th layer, and $\mathbf{f}^{l}_{ij}$ denotes the input feature vector of its $j$-th nearest neighbor $m_{ij}\in\mathcal{N}(i)$.
We first calibrate local information for each center, by learning an updated nearest-neighbor representation $\hat{\mathbf{f}}^{l}_{ij}\in \mathbb{R}^k$ in terms of $\mathbf{f}^{l}_{ij}$ and $\mathbf{f}^{l}_{i}$, as:

\begin{equation}
\label{update}
\hat{\mathbf{f}}^{l}_{ij}=MLP^{l}\Bigl(\mathbf{f}^{l}_{i} \oplus \mathbf{f}^{l}_{ij}\Bigl),\,\, \forall\, m_{ij}\in \mathcal{N}(i),
\end{equation}
where $\oplus$ indicates the channel-wise concatenation.
In this way, the information provided by $m_{ij}$ (encoded in $\hat{\mathbf{f}}^{l}_{ij}$) can be more consistent with the specific central node $m_i$, given the fact that $m_{ij}$ could be a nearest neighbor of more than one center, i.e., $\mathbf{f}^{l}_{ij}$ might be shared by multiple centers.

Additionally, we adopt a graph attention mechanism to aggregate the calibrated neighborhood information to each center. 
Inspired by \cite{Att1,Att2}, we choose a learning-based approach to estimate the attention weights for different neighbors. 
Compared with the strategy of using predefined weights~\cite{Att3}, learning the weights in a task-oriented fashion (e.g., by a lightweight network) can more flexibly capture local geometric characteristics of the dental model for the segmentation task. Specifically, we compute the attention weight $\mathbf{\alpha}^{l}_{ij} \in \mathbb{R}^k$ of neighbor $m_{ij}$ in the $l$-th layer as:
\begin{equation}
\label{weight}
\mathbf{\alpha}^{l}_{ij}=\sigma\Bigl(\Delta \mathbf{f}^{l}_{ij}\oplus \mathbf{f}^{l}_{ij}\Bigl), \,\,\forall\, m_{ij}\in \mathcal{N}(i),
\end{equation}
where the function $\sigma(\cdot)$ is implemented as a MLP in this work. It adopts both $\Delta \mathbf{f}^{l}_{ij} = \mathbf{f}^{l}_i-\mathbf{f}^{l}_{ij}$ and $\mathbf{f}^{l}_{ij}$ as the input, where $\Delta \mathbf{f}^{l}_{ij}$ quantifies the dissimilarity between $m_{i,j}$ and $m_i$ while $\mathbf{f}^{l}_{ij}$ provides detailed neighbor information in the feature space. 
Finally, the feature aggregation in the $l$-th layer is formulated as:
\begin{equation}
\label{aggregate}
\mathbf{f}^{l+1}_{i} = \sum_{m_{ij}\in \mathcal{N}(i)} \mathbf{\alpha}^{l}_{ij} \odot \hat{\mathbf{f}}^{l}_{ij},
\end{equation}
where $\mathbf{f}^{l+1}_{i}$ indicates the updated feature of center $m_i$, i.e., the input feature of the ($l$+1)-th layer. In Eq. (\ref{aggregate}), $\mathbf{\alpha}^{l}_{ij}$ and $\hat{\mathbf{f}}^{l}_{ij}$ are defined by Eq. (\ref{weight}) and Eq. (\ref{update}), respectively, and $\odot$ performs the element-wise production of two feature vectors. 

\paragraph{N-Stream.} Although the C-stream can learn the basic structure of a dental model from the 3D coordinates, it cannot sensitively distinguish between adjacent cells belonging to different classes (e.g., boundaries of teeth). 
Therefore, as complementary to the C-stream for accurate teeth delineation, we further design a N-stream to extract fine-grained boundary representations from the aspect of normal vectors in local areas.

Our N-stream takes the $M\times12$ matrix of normal vectors as input and consists of a series of graph max-pooling layers.
Notably, we force each layer in the N-steam to share the same KNN graph with the respective layer in the C-stream.
In this way, the graph max-pooling layers can focus on the learning of boundary representations in local regions, thereby avoiding the disturbance of distant cells that have similar normal vectors (but belonging to different classes).
For simplicity, we still use the symbols $\mathbf{f}^l_i$ and $\mathbf{f}^l_{ij}$ to denote the input features of a center node $m_i$ and its neighbor $m_{ij}$, respectively. Similar to the C-stream, the $l$-th layer of the N-stream first uses a MLP to learn the calibrated feature representation $\hat{\mathbf{f}}^{l}_{ij}$ for each $m_{i,j}$, i.e., similar to Eq.~(\ref{update}). Thereafter, we apply the channel-wise max-pooling on all neighbors' calibrated features to produce the boundary representation for the respective center, which can be formulated as:
\begin{equation}
\label{maxpooling}
\mathbf{f}^{l+1}_{i} = maxpooling\Bigl\{\hat{\mathbf{f}}^{l}_{ij},\,\, \forall \,m_{ij}\in \mathcal{N}(i)\Bigl\}.
\end{equation}
It is worth mentioning that we use max-pooling (rather than graph attention) in the N-steam since the max operator can sensitively capture the most distinctive features presented at the tooth boundaries.

\subsection{Feature Fusion}
\label{Feature FUsion}
Considering that the C-stream and the N-stream learn completely different feature representations from two complementary views, fusing their outputs can enable the overall network to comprehensively understand the structure of a dental model. 
To this end, as shown in Fig.~\ref{structure}, for each stream, we first use skip connections to concatenate its multi-scale cell-wise features from different layers (i.e., $\mathbf{F}^{l}_\mathbf{c}$ or $\mathbf{F}^{l}_\mathbf{n}$, where $l$ denotes the $l$-th layer), yielding a hierarchical feature matrix encoding local-to-global information. 
A MLP (i.e., MLP$\mathbf{_c}$ or MLP$\mathbf{_n}$) is then applied on this feature matrix to learn higher-level representations (i.e., $\mathbf{F_{c}}$ or $\mathbf{F_{n}}$) for the corresponding view (i.e, the C-stream or the N-stream), which can be formulated as follows:
\begin{equation}
\label{skip1}
\mathbf{F_c}=MLP_\mathbf{c}\Bigl(\mathbf{F}^{1}_\mathbf{c} \oplus \mathbf{F}^{2}_\mathbf{c} \oplus \mathbf{F}^{2}_\mathbf{c}\Bigl),
\end{equation}

\begin{equation}
\label{skip2}
\mathbf{F_n}=MLP_\mathbf{n}\Bigl(\mathbf{F}^{1}_\mathbf{n} \oplus \mathbf{F}^{2}_\mathbf{n} \oplus \mathbf{F}^{2}_\mathbf{n}\Bigl).
\end{equation}

Finally, the feature matrices from two complementary views are concatenated, which is followed by another MLP (i.e., MLP$\mathbf{_{pred}}$) to output an $M\times C$ matrix $\mathbf{P}$, with each row denoting the probabilities of a specific cell belonging to $C$ different classes, which can be formulated as: 
\begin{equation}
\label{predict}
\mathbf{P} = MLP_\mathbf{pred}\Bigl( \mathbf{F_c} \oplus \mathbf{F_n} \Bigl).
\end{equation}
We train TSGCNet with cross-entropy segmentation loss, which can be formulated as:
\begin{equation}
\label{loss}
loss = -\sum^{M}_{i=1}\sum^{C}_{c=1} p_{ic} \,{\rm log } \,y_{ic},
\end{equation}
where $p_{ic}$ and $y_{ic}$ denote the predicted and the ground-truth labeling probability for $c$-th class, respectively.

\subsection{Implementation Details}
\paragraph{Network Details.}
As shown in Fig.~\ref{structure}, the TSGCNet architecture consists of a C-stream, a N-stream, and a feature-fusion part. 
For each branch of the two streams, the MLPs in the first to the third layer contain one 1D Conv with 64 channels, 128 channels, and 256 channels, respectively.
The number $K$ of each KNN graph is set as 32. 
We use MLP to implement the graph attention function $\sigma(\cdot)$, which is followed by the channel-wise softmax to normalize the output weights.
In the feature fusion part, both MLP$\mathbf{_c}$ and MLP$\mathbf{_n}$ contain a 1D Conv with 512 channels, and MLP$\mathbf{_{pred}}$ contains four successive 1D Convs, each with 512, 256, 128, and $C$ channels, respectively. 
All 1D Convs are followed by batch normalization and LeakyReLU, except the last one in MLP$\mathbf{_{pred}}$, which is followed by a tensor-reshape operation to output the $M \times C$ probability matrix.
\paragraph{Training Details.}
Our TSGCNet was trained by minimizing the cross-entropy segmentation loss on two NVIDIA GTX 1080 GPUs for 200 epochs. We use the Adam optimizer with the mini-batch size setting as 4. The initial learning rate was 1e-3, which was reduced by 0.5 decay for every 20 epochs. 

\begin{figure}[H]
\begin{center}
\includegraphics[scale=0.6]{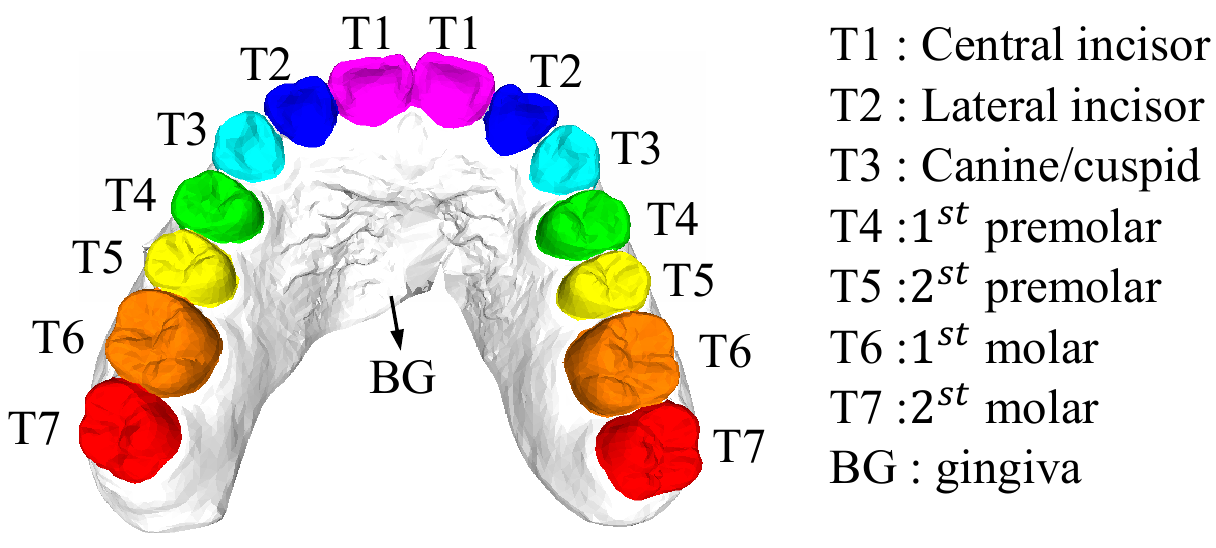}
\end{center}
   \caption{An illustration of a manually labeled 3D dental model. Each dental model includes 8 classes, i.e., the symmetric central incisor, lateral incisor, canine, $1^{st}$ premolar, $2^{st}$ premolar, $1^{st}$ molar, $2^{st}$ molar, and the gingiva.}
\label{label}
\vspace{-1.5em}
\end{figure}

\begin{table*}[t]
\caption{The segmentation results for five competing methods and our method on OA and mIoU.}
\begin{center}
\begin{tabular}{ccccccccccc}
\hline
Method     & OA             & mIoU           & T1             & T2             & T3             & T4             & T5             & T6             & T7             & BG             \\ \hline
PointNet\cite{Pointnet}   & 84.95          & 66.86          & 55.31          & 65.31          & 69.35          & 75.47          & 72.21          & 66.18          & 74.71          & 84.86          \\
PointCNN\cite{pointcnn}   & 88.61          & 72.86          & 61.72          & 66.45          & 68.10          & 78.98          & 78.57          & 70.51          & 72.15          & 86.39          \\
PointNet++\cite{Pointnet++} & 90.25          & 78.14          & 67.82          & 74.61          & 78.10          & 82.73          & 80.70          & 74.67          & 78.94          & 87.52          \\
DGCNN\cite{spatial1}      & 91.93         & 84.30          & 82.18         & 79.95          & 82.09          & 87.88          & 86.24          & 80.14          & 84.26          & 91.65          \\
MeshSegNet\cite{dental4}     & 93.11          & 84.47          & 81.31          & 83.65          & 82.15          & 82.87          & 84.81          & 81.93          & 87.10          & 91.94          \\
Ours       & \textbf{95.25} & \textbf{88.99} & \textbf{86.01} & \textbf{87.48} & \textbf{89.38} & \textbf{90.44} & \textbf{89.54} & \textbf{85.99} & \textbf{89.32} & \textbf{93.76} \\ \hline
\end{tabular}
\end{center}
\label{result}
\vspace{-1em}
\end{table*}

\section{Experimental Results}
\subsection{Dataset}
The studied dataset consists of 80 3D dental models acquired by an IOS for different orthodontic patients. Each raw dental model roughly contains more than 100,000 meshes, which were downsampled to 16,000 (e.g. $M=16,000$) meshes through the reduction of redundant information, while preserving the original topology. The dataset was randomly split as a training set with 64 subjects, and a testing set with 16 subjects. Our target is to automatically segment each dental model as $C=8$ different semantic parts, including the central incisor (T1), lateral incisor (T2), canine/cuspid (T3), 1$^{\text{st}}$ premolar (T4), 2$^{\text{nd}}$ premolar (T5), 1$^{\text{st}}$ molar (T6), 2$^{\text{nd}}$ molar (T7), and background/gingiva (BG). The ground-truth annotations of all dental models were defined by an experienced orthodontist, with a typical example shown in Fig.~\ref{label}.

\subsection{Experimental Setup}
\paragraph{Data Augmentation.} We augment the training set by the combination of 1) random translation, and 2) random rotation of each 3D dental model. 
Specifically, each training dental model is translated with a displacement randomly sampled between $[-10,10]$ and rotated along the $y$-axis with an angle randomly sampled between $[-\frac{\pi}{6}, \frac{\pi}{6}]$. 
In this way, we generate 64 new samples from the original dental models to enrich the diversity of the training set.

\paragraph{Competing Methods.}
Our TSGCNet was compared with five state-of-the-art methods for both 3D shape segmentation (i.e., PointNet~\cite{Pointnet}, PointNet++~\cite{Pointnet++}, PointCNN~\cite{pointcnn}, DGCNN~\cite{spatial1}) and 3D dental model segmentation (i.e., MeshSegNet~\cite{dental4}). 
All these five competing methods were implemented by their original codes and trained on the same dataset. 
For the grouping operations of PointNet++~\cite{Pointnet++}, we deployed the 3D coordinates of the central point of each cell to compute the spatial distance.
The overall segmentation performance (averaged over all classes) was quantitatively evaluated by two metrics, i.e., 1) Overall Accuracy (OA), and 2) mean Intersection-over-Union (mIoU). Besides, we also quantify the detailed IoU of each class.

\subsection{Comparison with Competing Methods}
The overall segmentation results are presented in Table \ref{result}. Results show that our method achieves the best performance in terms of both OA and mIoU metrics. In particular, when compared with the competing method in this specific task, MeshSegNet \cite{Att2}, which directly consumes the combination of coordinates and normal vectors, the proposed TSGCNet still increases the segmentation accuracy by $2.14\%$ and $4.52\%$ on the OA and mIoU, respectively. Additionally, our method also significantly outperforms the graph based network DGCNN~\cite{spatial1}, demonstrating the effectiveness of the proposed two-stream mechanism that can learn more discriminative geometric feature representations for accurate tooth segmentation.
Furthermore, despite the varying shape appearances of different types of teeth, our method is able to present consistent superior segmentation performance over other approaches by a large margin.

\begin{figure*}[t]
\begin{center}
\includegraphics[scale=0.94]{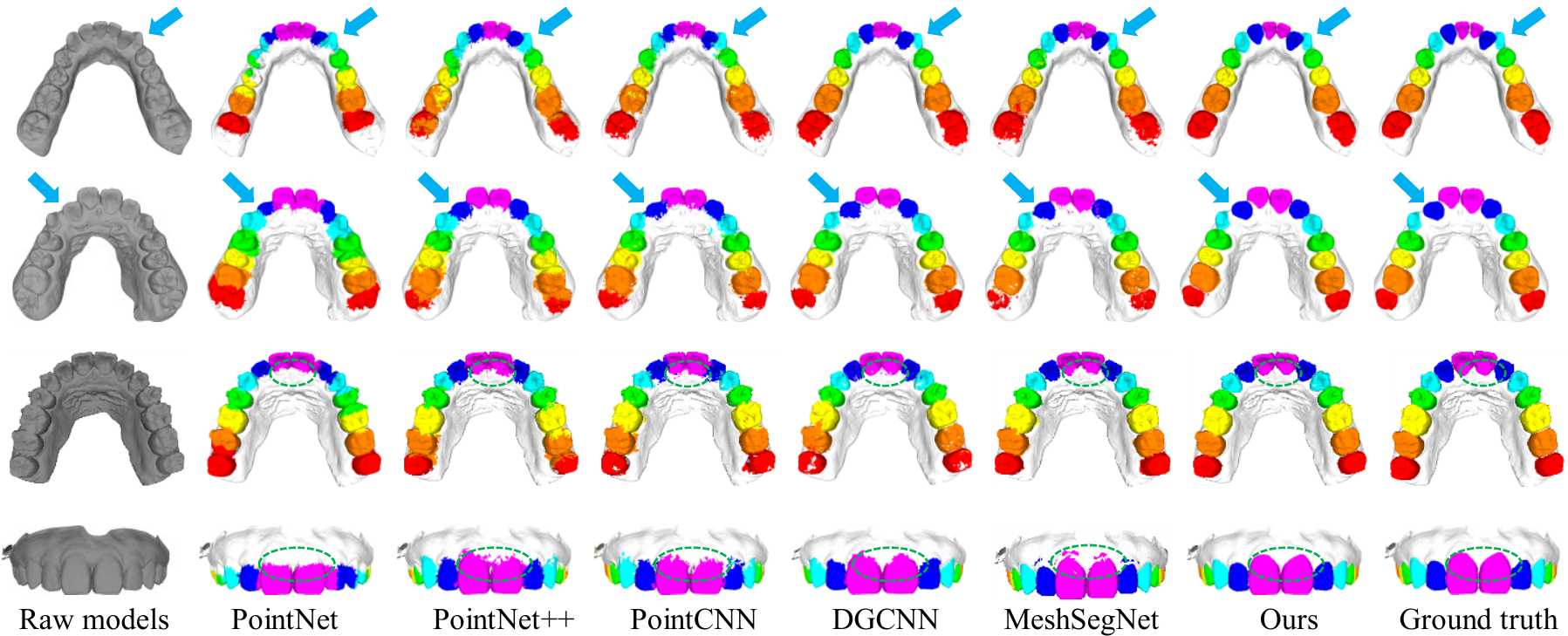}
\end{center}
   \caption{Visualization of representative segmentation results produced by five competing methods and our method, along with the respective ground-truth annotations.}
\label{segmental result}
\end{figure*}

We also visualize the segmentation results (obtained by different methods) for four representative dental models in Fig.~\ref{segmental result}. 
In consistency with the quantitative evaluations, we can observe from Fig.~\ref{segmental result} that our TSGCNet also qualitatively outperforms all the competing methods, especially for the challenging areas marked by the blue arrows and green dotted circles. 
Specifically, in the area of teeth misalignment (indicated by the blue arrows in the first two rows), PointNet~\cite{Pointnet}, PointNet++~\cite{Pointnet++} and PointCNN~\cite{pointcnn} either result in under-segmentation or over-segmentation.
Graph-based competing methods (i.e., DGCNN~\cite{spatial1} and MeshSegNet~\cite{dental4}) achieve better performance based on the extraction of detailed local spatial information. However, they still fail to capture the complete teeth structure. 
In contrast, due to the use of complementary information from the C-stream and N-stream, our TSGCNet achieves more accurate results than all the competing methods in these misaligned areas. 
Besides, from the third and fourth rows of Fig.~\ref{segmental result}, we can see that our proposed method can also better distinguish the boundaries between adjacent teeth, especially for the two adjacent incisors (indicated by the green dotted circles). 
Finally, when comparing our method with MeshSegNet~\cite{dental4} in the fourth row, we can see that MeshSegNet~\cite{dental4} produces many isolated false predictions on gingiva, even those mislabeled mesh cells are relatively far away from the real tooth area. 
This further suggests that the direct concatenation of normal vectors and coordinates as a single feature vector (e.g., in MeshSegNet) may confuse the learning of discriminative geometric features in some cases, while the two-stream structure (i.e., in our TSGCNet) is a more appropriate design.

\section{Ablation Study}
In this section, we conduct detailed ablation studies to evaluate the efficacy of the critical components of our TSGCNet. 

\subsection{Effectiveness of the Two-Stream Structure}
\label{5.1}
In this series of experiments, we first evaluate the effectiveness of our two-stream structure.
Specifically, we remove the N-stream (i.e., only adopting the C-stream with the coordinates as input) or the C-stream (i.e., only adopting the N-stream with the normal vectors as input) to generate two different variants of TSGCNet, which are denoted as \textbf{TSGCNet-C} and \textbf{TSGCNet-N}, respectively. In addition, we also build another single-stream variant of TSCGNet (denoted as \textbf{TSGCNet-S}) that directly learns from the combination of the coordinates and normal vectors. 
Note that TSGCNet-S has a similar structure to TSGCNet-C but with different input. 
We compare these three variants with the final TSGCNet, with the quantitative results listed in Table \ref{ablation1}. 
It can been seen that TSGCNet-N and TSGCNet-C led to worse results than both TSGCNet-S and TSGCNet. This justifies the complementarity between the geometric information provided by the coordinates and normal vectors in delineating teeth on dental models.
On the other hand, when compared with TSGCNet-S, the original TSGCNet further improves the segmentation accuracy, which suggests the effectiveness of our two-stream structure in extracting the complementary geometric information from the two different views. 

\begin{table}[H]
\caption{The segmentation results for the original TSGCNet and three variants. TSGCNet-C and TSGCNet-N stand for the sole use of the C-stream and N-stream, respectively. TSGCNet-S denotes the single-stream version of TSGCNet, which directly concatenates the coordinates and normal vectors as input.}
\begin{center}
\begin{tabular}{lll}
\hline
Structure           & OA    & mIoU  \\ \hline
TSGCNet-C           & 83.23 & 63.79 \\
TSGCNet-N           & 55.42 & 20.77 \\
TSGCNet-S           & 87.25 & 73.44 \\
TSGCNet             & \textbf{95.25} & \textbf{88.99} \\ \hline
\end{tabular}
\end{center}
\label{ablation1}
\vspace{-1em}
\end{table}

\subsection{Effectiveness of Feature-Aggregation Strategy}

As described in Section \ref{Two Stream}, we use two different feature aggregation strategies in the C-stream and N-stream of our TSGCNet. Specifically, the graph attention aggregation is used in the C-stream, while the graph max-pooling aggregation in the N-stream. 
To evaluate the effectiveness of our design, we implement three variants of TSGCNet by changing the feature aggregation strategy in each stream, including \textbf{1)} both streams use max-pooling, \textbf{2)} both streams use attention, and \textbf{3)} C-stream uses max-pooling while N-stream uses attention. For simplicity, we denote those three variants and the original TSGCNet as \textbf{M+M}, \textbf{A+A}, \textbf{M+A}, and \textbf{A+M}, respectively. We then compare the segmentation results of these variants in Table \ref{ablation2}. From Table \ref{ablation2}, we can see that using attention mechanisms in the C-stream can achieve better performance (please refers to A+M  \emph{vs}. M+M) when compares with the use of max-pooling, which suggests that graph attention aggregation can extract finer details of the tooth shape from coordinates. Besides, using max-pooling in the N-stream can further refine the segmentation results (please refers to A+M \emph{vs}. A+A). This can be rationally explained by that max-pooling can extract more distinctive morphological features, which in return helps the network to capture difference between neighboring cells, especially at the tooth boundaries.
\begin{table}[H]
\caption{The segmentation results by using different feature aggregation strategies. M+M (or A+A) stands for using max-pooling (or attention) in both two streams. M+A stands for using max-pooling and attention in the C-stream and N-stream, respectively. A+M denotes the original TSGCNet.}
\begin{center}
\begin{tabular}{lll}
\hline
Structure           & OA    & mIoU  \\ \hline
M+M           & 94.56 & 86.24 \\
A+A           & 95.01 & 87.35 \\
M+A           & 93.93 & 85.67 \\
A+M             & \textbf{95.25} & \textbf{88.99}\\ 
\hline
\end{tabular}
\end{center}
\label{ablation2}
\end{table}

\begin{figure}[H]
\begin{center}
\includegraphics[scale=0.61]{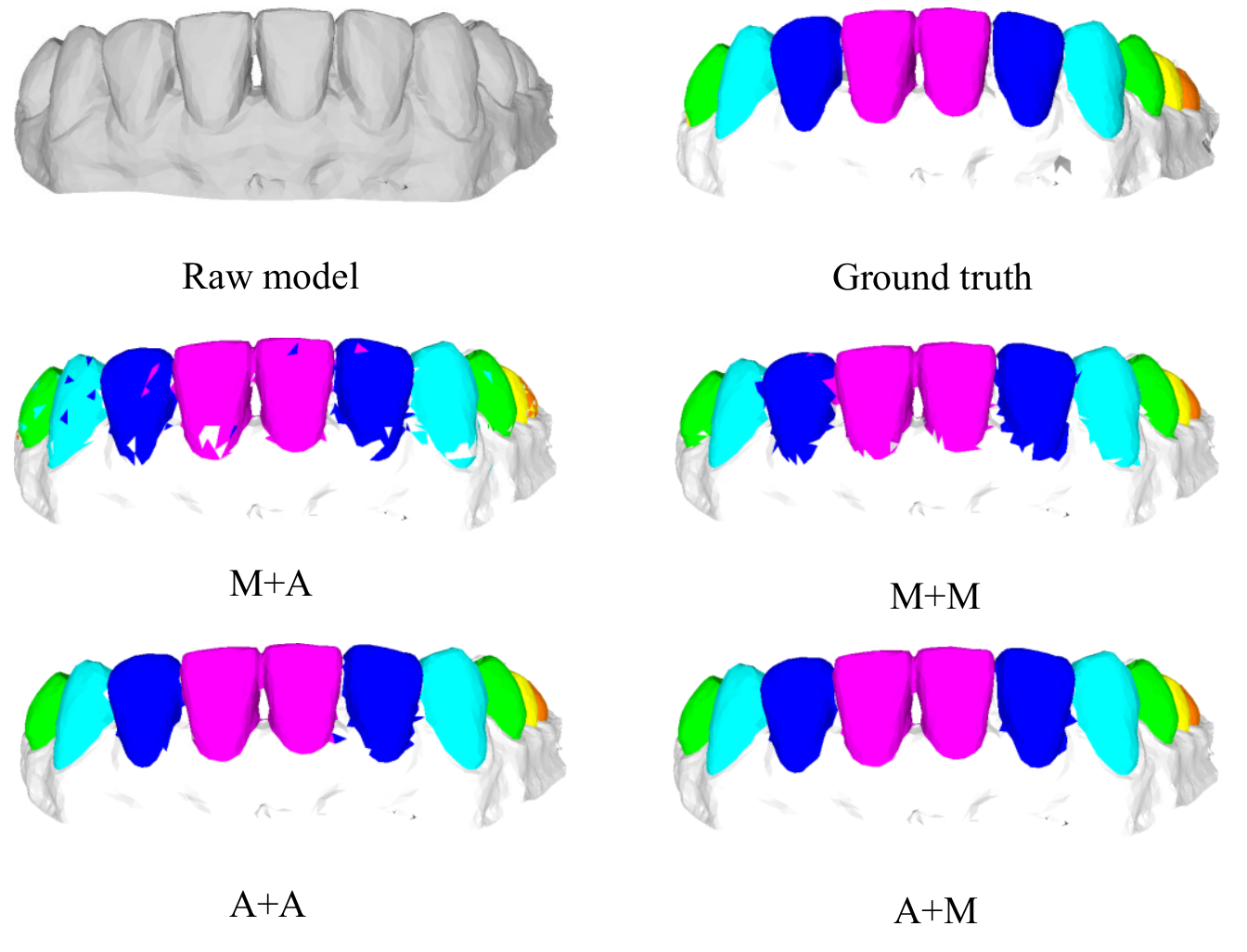}
\end{center}
   \caption{A segmentation example for TSGCNet by using different feature aggregation strategies.}
\label{aggregation_example}
\end{figure}

We also show the segmentation results of a typical example obtained by these variants in Fig.~\ref{aggregation_example}.
In consistency with quantitative evaluations in Table \ref{ablation2}, we can see that both M+A and M+M have more outliers than A+A and A+M, which further confirms that graph attention aggregation is more suitable for the C-stream. Besides, when comparing A+A with A+M, we also observe that A+M generates more precise segmentation on boundaries, which further confirms that graph max-pooling aggregation is more suitable for the N-stream.

\subsection{Effectiveness of Feature-Fusion Strategy}
As described in Section~\ref{Feature FUsion}, the multi-scale high-level feature produced by C-stream and N-stream (i.e., $\mathbf{F_{c}}$ and $\mathbf{F_{n}}$) are fused to learn complementary information in our TSGCNet. To evaluate the effectiveness of this high-level feature fusion strategy, we compare the TSGCNet with another variant that is implemented by applying a low-level feature fusion strategy. Specifically, during the two-stream feature extraction stage, the output of the $l$-th layer in both streams are concatenated (i.e., $\mathbf{F^{l}_{c}}$ and $\mathbf{F^{l}_{n}}$ are concatenated) as the input of the $(l+1)$-th layer. This means that the C-stream and N-stream have the same input in the $(l+1)$-th layer. We denote our original feature fusion strategy and the variant as \textbf{H-fusion} and \textbf{L-fusion}, respectively. 

\begin{table}[H]
\caption{The segmentation results for two different feature fusion strategies. The L-fusion denotes low-level feature fusion strategy, and the H-fusion stands for our adopted feature fusion strategy.}
\begin{center}
\begin{tabular}{lll}
\hline
Strategy           & OA    & mIoU  \\ \hline
L-fusion           & 93.28 & 85.49 \\
H-fusion             & \textbf{95.25} & \textbf{88.99} \\ \hline
\end{tabular}
\end{center}
\label{ablation3}
\vspace{-1em}
\end{table}

We further compare the segmentation results of H-fusion and L-fusion, as shown in Table \ref{ablation3}. From this table, it can be seen that the OA and mIoU of H-fusion is 1.97\% and 3.50\% higher than L-fusion, respectively. It is potentially because the premature feature fusion also confuses the learning of discriminative features. Besides, due to different properties between coordinates and normal vectors, the KNN graph built on the concatenated features may result in a random distribution of neighbors in real space, which tends to hamper the network to learn local-to-global information.

\subsection{Limitations}
\begin{figure}[H]
\begin{center}
\includegraphics[scale=0.73]{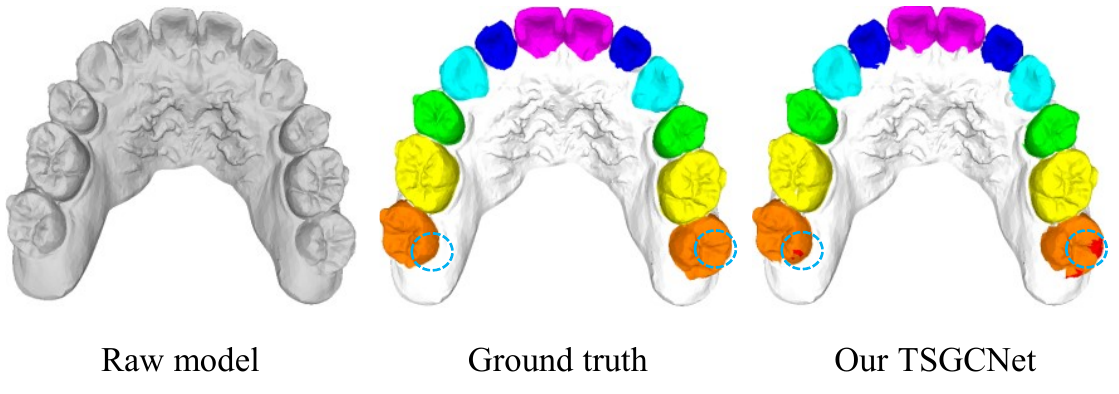}
\end{center}
\caption{A segmentation example for the 12-teeth dental model produced by our TSGCNet.}
\label{limitation}
\end{figure}

Although our TSGCNet has achieved the leading performance in the task of 3D dental segmentation, it still has certain limitations. Most typically, TSGCNet cannot robustly handle special cases with 12 teeth. For example, we showed the segmentation result of one dental model with 12 teeth in Fig.~\ref{limitation}, which can be seen that our TSGCNet generates many false predictions on T6 (indicated by the blue dotted circles). This can be possibly interpreted by the fact that the outermost tooth of 12-teeth dental models is annotated as T6, which is usually annotated as T7 in the normal dental models. To address this problem, including more 12-teeth cases as training samples would be considered in our future research.

\section{Conclusion}
A two-stream network, called TSGCNet, has been proposed in this paper to automatically segment individual tooth from 3D dental models acquired by intra-oral scanners. To eliminate the mutual confusion caused by mixed geometric inputs, the proposed TSGCNet apply two input-aware graph-learning streams to independently extract discriminative geometric features from coordinates and normal vectors, respectively. Feature representations produced by two different stream are then fused to learn complementary multi-view information for the end-to-end cell-wise prediction. An extensive comparison has been performed between our TSGCNet and other five state-of-the-art methods on a real-patient dataset, and the corresponding results demonstrate the superiority of our proposed method, especially for practically challenging cases.

{\small
\bibliographystyle{ieee_fullname.bst}
\bibliography{egbib}
}

\end{document}